\documentclass[runningheads]{llncs}
\usepackage[T1]{fontenc}
\usepackage{booktabs} 
\usepackage{amsmath}
\usepackage{makecell}
\usepackage{graphicx}
\usepackage{xcolor}
\usepackage{amssymb}
\usepackage{subcaption}
\usepackage{bbding}
\usepackage[misc]{ifsym}
\begin{document}

\title{SCRA-VQA: Summarized Caption-Rerank for Augmented Large Language Models in Visual Question Answering}
\titlerunning{Summarized Caption-Rerank for Augmented VQA} 
%
%
\author{Yan Zhang\inst{1,2,3} \and
Jiaqing Lin\inst{1} \and
Miao Zhang\inst{1,2,3}\and
Kui Xiao\inst{1,2,3}\and
Xiaoju Hou\inst{4}\and \\
Yue Zhao\inst{5}\and
Zhifei Li\inst{1,2,3}\textsuperscript{(\Letter)}}

%
\authorrunning{Y. Zhang et al.}
\institute{School of Computer Science, Hubei University, Wuhan, China 
\email{zhifei1993@hubu.edu.cn} \and Hubei Key Laboratory of Big Data Intelligent Analysis and Application (Hubei University), Wuhan, China \and Key Laboratory of Intelligent Sensing System and Security (Hubei University), Ministry of Education, Wuhan, China \and Institute of Vocational Education, Guangdong Industry Polytechnic University, Guangzhou, China \and Shandong Police College, Jinan, China
}
\maketitle           
\pagenumbering{gobble}
\begin{abstract}
\sloppy
Acquiring high-quality knowledge is a central focus in Knowledge-Based Visual Question Answering (KB-VQA).  Recent methods use large language models (LLMs) as knowledge engines for answering. These methods generally employ image captions as visual text descriptions to assist LLMs in interpreting images. However, the captions frequently include excessive noise irrelevant to the question, and LLMs generally do not comprehend VQA tasks, limiting their reasoning capabilities. To address this issue, we propose the Summarized Caption-Rerank Augmented VQA (SCRA-VQA), which employs a pre-trained visual language model to convert images into captions. Moreover, SCRA-VQA generates contextual examples for the captions while simultaneously summarizing and reordering them to exclude unrelated information. The caption-rerank process enables LLMs to understand the image information and questions better, thus enhancing the model's reasoning ability and task adaptability without expensive end-to-end training. Based on an LLM with 6.7B parameters, SCRA-VQA performs excellently on two challenging knowledge-based VQA datasets: OK-VQA and A-OKVQA, achieving accuracies of 38.8\% and 34.6\%. Our code is available at \url{https://github.com/HubuKG/SCRA-VQA}.
\end{abstract}
\section{Introduction}
Visual Question Answering (VQA) is a crucial task that combines natural language processing and computer vision. Over the past few years, substantial progress has been made in the field of VQA \cite{1,2}. Among the VQA tasks, knowledge-based visual question answering (KB-VQA) is particularly challenging, requiring external knowledge beyond the given question and visual content to provide accurate answers. To solve such a task, a model must possess strong visual perception and reasoning abilities, along with the capacity to effectively integrate world knowledge from external knowledge bases (e.g., Wikipedia). Therefore, accurately acquiring the required knowledge has become the core problem of KB-VQA.

In the current study, researchers focus on image recognition and knowledge-based logical reasoning \cite{3}. Image recognition aims at obtaining more visual information to help understand user questions \cite{4}. Logical reasoning aims to identify associations between knowledge to reason the answer accurately \cite{6}. Simultaneously, large language models (LLMs) like GPT-3 have demonstrated excellent capabilities in knowledge retrieval and question-answering tasks in natural language processing. Inspired by these LLMs, Prophet \cite{7} achieves better results on knowledge-based VQA tasks by prompting GPT-3 for reasoning, outperforming traditional large vocabulary models. Repurposing pre-trained LLMs for new tasks is ideal from a resource perspective because it avoids the high computational costs of traditional model-building methods.

\begin{figure}[t]
\centering
\includegraphics[scale=0.8]{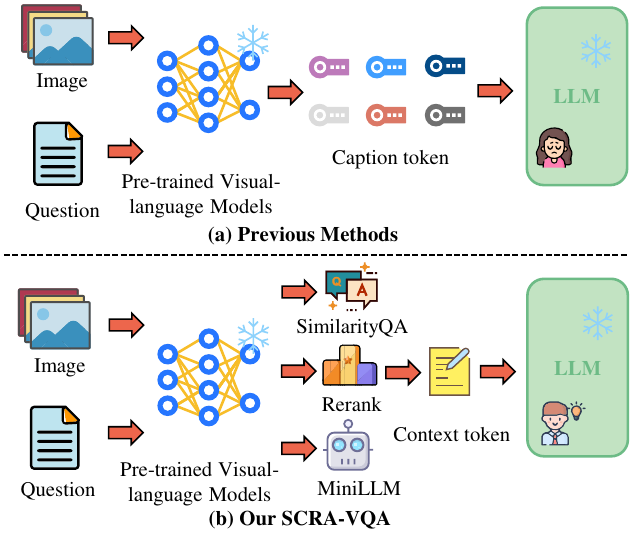} 
\caption{Comparison between existing prompting methods and SCRA-VQA on VQA tasks using frozen LLMs. Unlike previous methods that provide image-to-text information directly into LLMs, SCRA-VQA processes information in three distinct ways before passing it on to help better understand the task.}
\label{figure:1}
\end{figure}

However, as the demand for solving more complex problems grows, the knowledge required for knowledge-based logical reasoning becomes increasingly complex. Meanwhile, the mismatch between the generated image description and the question may make the LLMs less powerful. For instance, a caption that says, “\textit{a group of puppies playing on the grass}” would be unhelpful for a question like “\textit{What season is depicted in the image?}” and might even hinder the LLMs from generating the correct answer. In addition, since LLMs only take in text, the precision of their logical reasoning is almost dependent on the acquired information. Relying only on the captions of a few relevant entities may lead to incorrect judgment and reduced performance. Thus, activating the performance of LLMs is a crucial issue.

To address these issues, this paper proposes a Summarized Caption-Rerank Augmented VQA (named SCRA-VQA) method based on frozen LLMs. As shown in Fig.~\ref{figure:1}, it demonstrates the VQA task and conveys the image's content to the LLM for answering questions. Compared to previous methods, our method uses a pre-trained vision-language model \cite{8} to generate captions for images. Then, a question generation model is used to convert the captions into similar answer-question pairs as contextual examples \cite{9}, enabling the LLM to understand the VQA task better. Furthermore, considering the large number of generated captions and the lack of connection between them, we summarized and Rerank it to enable LLMs to understand image information while reducing data processing stress fully. 

In this paper, we have made the following contributions:
\begin{itemize}
\item We propose a Summarized Caption-Rank Augmented VQA (SCRA-VQA) method, a new method for processing captions to assist LLM in VQA tasks. This method ensures that captions are only relevant to the information needed for the question, enabling better connections between vision and language modalities.
\item SCRA-VQA enables pre-trained LLMs to execute zero-shot VQA without requiring costly end-to-end training or specialized text networks. This method facilitates low-cost and adaptable model deployment, providing an effective and lightweight solution.
\item Our experimental results demonstrate that the OPT model effectively achiev-es zero-shot VQA when equipped with SCRA-VQA. Moreover, our method outperforms the state-of-the-art Img2Prompt methodology on the A-OKVQA dataset by 3.9\%. 
\end{itemize}

\section{Related Work}
\label{sec:2}
\subsection{Visual Question Answering}
VQA has always been a central focus of research. Recent VQA work mainly focuses on two areas: 1) better visual feature extraction and 2) more advanced model architectures.

Visual feature extraction has always been a key research direction in VQA. Numerous methods continuously explore how to more effectively utilize visual information to improve the performance of question-answering systems. For instance, the bottom-up and top-down attention mechanism \cite{12} allows attention to be calculated at the level of visual objects. CLIP-ViL \cite{13} introduces a novel training methodology using a contrastive loss function to learn visual concepts directly from natural language descriptions. This enables zero-shot solid performance across a broad range of visual tasks without requiring task-specific training. VinVL \cite{14} improves visual representations in vision-language tasks by developing an enhanced object detection model pre-trained on extensive datasets. Overall, these innovative methods collectively enhance the effectiveness and robustness of visual feature extraction in VQA tasks.

Recent VQA frameworks have made substantial progress by integrating visual and linguistic information. Among these, BLIP \cite{8} introduces a vision-language pre-training framework that enhances performance by generating synthetic captions and filtering noise from web-derived data. N2NMNs \cite{15} develop End-to-End Module Networks that autonomously generate dynamic layouts from textual queries for efficient visual question answering. Additionally, certain studies have concentrated on neurosymbolic VQA, which involves deriving a structural scene representation from an image and generating a program sequence based on the question. In particular, VQs \cite{16} gather visual questions and segmentation answers, establishing a connection between instance segmentation and the question-answer pairs in the VQA dataset. NS-VQA \cite{17} introduces a model that identifies scene objects and employs a symbolic program executor for reasoning and responding to inquiries. DCR \cite{18} suggests a profound concept reasoner to create syntactic rule structures from concepts and apply these rules to meaningful concepts for interpretable predictions. These innovative methods collectively enhance the accuracy and interpretability of the visual-to-text process in VQA tasks. Based on these methods, our method demonstrates that image-to-text transforms and the critical information of the question can complement each other. This synergy provides more complete visual information, which leads to improved knowledge retrieval and VQA performance.

\subsection{Knowledge-based Visual Question Answering}
KB-VQA requires combining images with questions to seek answers from external knowledge sources \cite{19}. Therefore, acquiring knowledge is a key focus of the KB-VQA task. There are two ways to acquire knowledge: explicit knowledge stored in knowledge bases and implicit knowledge embedded in LLMs.

Extracting accurate knowledge from knowledge bases has always been a focus of research.  To extract knowledge from the knowledge base, Hypergraph Transformer \cite{6} first identifies keywords in the questions. Then, these keywords are used to search for corresponding knowledge in the knowledge base. Other methods convert images into text labels for retrieval. However, this often results in the inclusion of irrelevant information. Thus, KAN \cite{20} utilizes keywords from the question to align with the retrieved knowledge, subsequently discarding any knowledge that doesn't match the question. Meanwhile, some methods use deep learning networks to filter knowledge. For instance, one method utilizes a Graph Neural Network (GNN) to embed images, questions, and knowledge facts simultaneously, thereby facilitating the selection of pertinent knowledge \cite{22}. KVQA \cite{23} integrates an attention module that assigns greater weights to knowledge pertinent to the question. These methods collectively enhance the precision and relevance of knowledge extraction from knowledge bases.

Unlike methods that depend on explicit knowledge to answer questions, an increasing number of methods are now trying to extract implicit knowledge from LLMs for KB-VQA. KAT \cite{24} uses LLMs to provide candidate answers, while PICa \cite{5} directly uses GPT-3 for answer prediction. KAPT \cite{25} also incorporates diverse knowledge into their prompt tuning method. This method boosts the zero-shot abilities of LLMs, enabling them to adapt to various downstream tasks. These methods generally employ a suitable information processing technique to prompt LLMs for VQA tasks. However, this also brings a new problem: LLMs tend to overlook task-relevant information when dealing with a large amount of textual information. Compared with the previous methods, SCRA-VQA alleviates this issue by evaluating the relevance of text generated from images and summarizing its content.

\begin{figure*}[!t]
\centering
\includegraphics[width=1.0\textwidth]{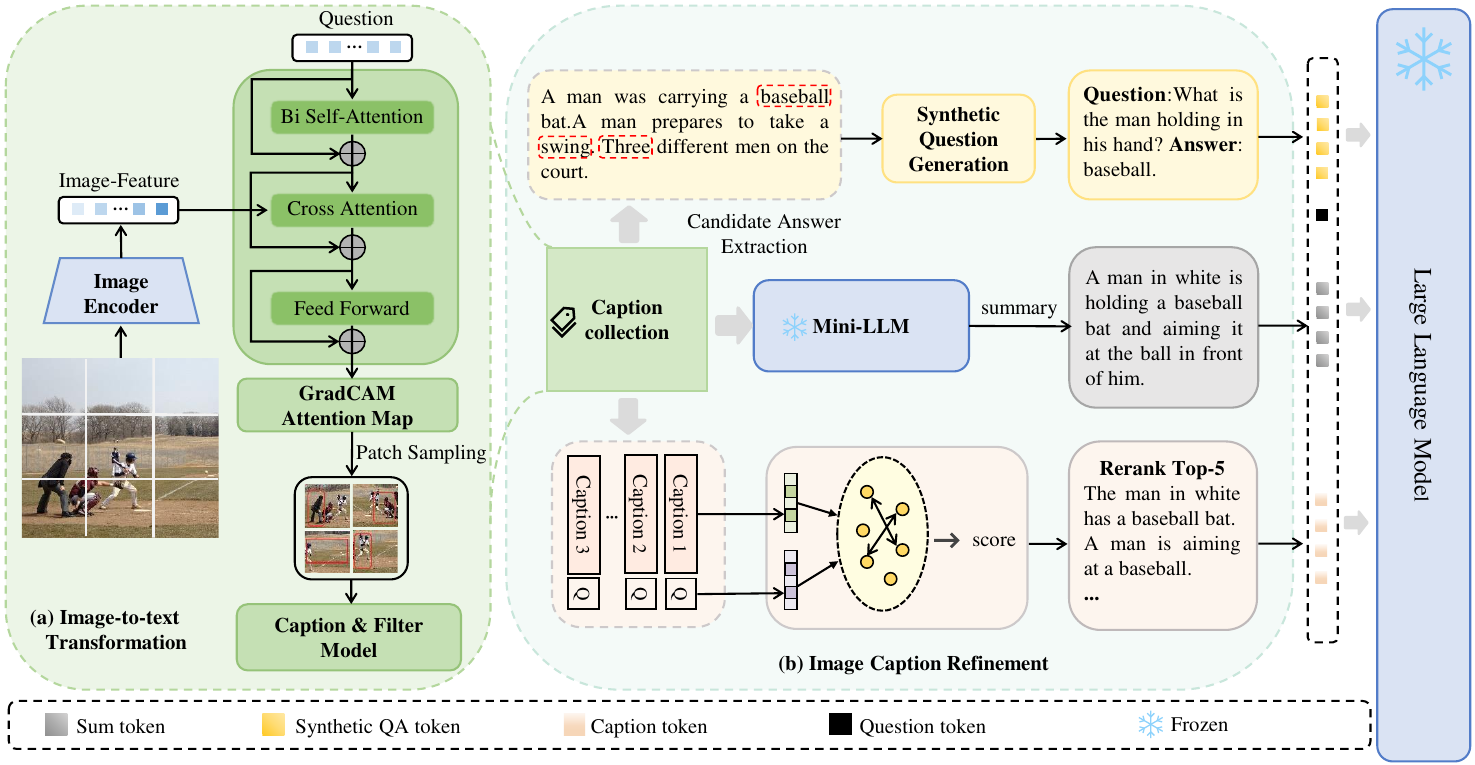}
\caption{The overall process of SCRA-VQA. (a) The Image-to-text Transformation module obtains the image caption that is most relevant to the question. (b) The image Caption Refinement module provides the generated caption to LLMs after similarity problem generation, Summary, and Rerank.}
\label{figure:2}
\end{figure*}

\section{Method}
\label{sec3}
\subsection{Image-to-text Transformation}
\label{sec:3.1}
Given that images often hold abundant information, the question might only pertain to specific regions of the image. As shown in Fig.~\ref{figure:2}, it pertains only to the man in white clothing, while the question is “\textit{What will the man in white do next?}”. However, the caption generated from the entire image will likely focus on other individuals, directly impacting the LLM’s ability to generate an answer.

To derive the cross-attention scores $W$ between each image region and each discussed token, inspired by Img2Promp \cite{10}, our method leverages the Image-Text Encoder (ITE) from BLIP \cite{8} for the computation:
\begin{equation}
W=\mathrm{softmax}\left(\frac{f_qW_Q(W_K)^\top(f_v)^\top}{\sqrt{D_q}}\right),
\end{equation}
where $W_{Q}$ is the query head and $W_{K}$ is the key head. $f_{v}$ and $f_{q}$ represent the features of the image region and the textual features, respectively. $D_{q}$ is the dimension of the textual feature. After obtaining the scores of each token and image block, we follow the GradCAM \cite{26} method to calculate the relevance score $R$ of each image block with the question:
\begin{equation}
    R=\frac{1}{G}\sum_{g=1}^G\sum_{l=1}^N\min\left(0,\frac{\partial\text{sim}(v,q)}{\partial W_{l}^g}\right)W_{l}^g,
\end{equation}
where sim($v$,$q$) is the cross-attention score of the ITE function. $G$ is the number of attention heads, and $N$ is the number of text tokens. After obtaining the patch relevance scores $R$, we perform TOP-\textit{K} sampling and generate descriptions for the resulting image blocks.

\begin{table}[t]
\centering
\caption{Templates for generating questions are provided by different parts of speech of words.}
\begin{tabular}{cc}
\toprule
Type      & Question   Templates \\  \midrule 
Noun      & \makecell[c]{What item is this in this picture? \\ What item is that in this picture?} \\ \midrule 
Verb      & \makecell[c]{What action is being taken in this picture? \\ Why is this item in this picture? \\ Which action is being taken in this picture?  \\ What action is the item doing in this picture?} \\ \midrule 
Adjective & \makecell[c]{How to describe one item in this picture?  \\ What is the item’s ADJ TYPE in this picture? \\ What is the ADJ   TYPE in this picture?}  \\
\bottomrule
\end{tabular}
\label{table:1}
\end{table}

\subsection{Image Caption Refinement}
\label{sec:3.2}
\subsubsection{Similarity Question Generation.}
Previous method \cite{6} has shown that providing GPT-3 with appropriate question-answer pairs as contextual examples can significantly improve performance. Building on this, we employ the VQ2A \cite{9} method to create question-answer pairs by using question templates to generate similar pairs. The question generation template comprises three steps: extracting candidate answers from captions using syntactic parsing, generating contextually consistent questions via a neural model, and filtering question-answer pairs through consistency validation. Table~\ref{table:1} shows the templates used in this paper for question generation. For example, when the part of speech is a noun, the template questions are “\textit{What item is this in this picture?}” and “\textit{What item is that in this picture?}”.

\subsubsection{Captions to Rerank.}
\begin{figure}[!t]
\centering
\includegraphics[width=0.8\textwidth]{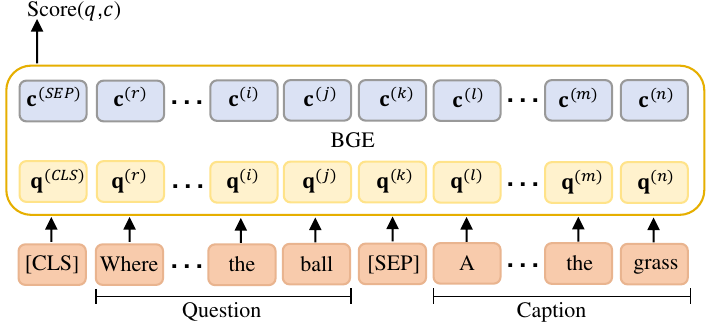} 
\caption{Rerank independently calculates the relevance score between a specific caption and a question.}
\label{figure:3}
\end{figure}

After successfully executing the image-to-text conversion process on pertinent image regions, we observed the reduction of irrelevant captions. However, their number is still large, which is still a challenge for LLMs. In response to this issue, our model introduces a novel one-to-one Rerank procedure from the field of Retrieval-Augmented Generation (RAG), as shown in Fig. \ref{figure:3}. This refined matching strategy enables a more precise calculation of text relevance by directly comparing individual pairs of captions and questions. This enhances the specificity and accuracy of information retrieval. The relevance score $s$ for the captions and the selection process are defined as follows:
\begin{equation}
s=\text{score}(q,c)=\mathbf{v}_{p}^{{\top}} cls \left(\text{BGE}\left(\text{concat}(q,c)\right)\right),
\end{equation}
where $cls$ extracts BGE’s [CLS] vector and $\mathbf{v}_{p}^{{\top}}$ is a projection vector. It samples query document pairs independently and computes each query-document pair. Once we have the scores for each caption, we sort them and select the top-5 with the highest scores as input captions:
\begin{equation}
 \mathcal{J}=\underset{j\{1,2,...,N\}}{\arg\text{TopK}}s_{[j]},
\end{equation}
where $s_{[j]}$ represents the set of captions following the scoring process. $\mathcal{J}$ represents the five captions selected with the highest scores.

\subsubsection{Gemma2b Caption Summarization.}
Although there is a wealth of information in the captions, we have found that the LLM occasionally struggles to completely grasp the underlying relationships between them. Meanwhile, Google's latest Gemma2b model offers robust performance without requiring significant computational resources. Thus, SCRA-VQA uses the Gemma2b model to summarize captions, distilling essential information into a format better suited for LLMs. This refinement enhances knowledge transfer, significantly improving answer prediction accuracy. The method boosts information processing efficiency, ensuring LLMs reliably acquire and use necessary knowledge for accurate answers.

\subsection{Prompt Design}
\label{sec:3.3}
Prompts play a crucial role in the inference process of LLMs, guiding them to fully utilize their understanding and reasoning capabilities. Therefore, we have developed a comprehensive and prompt design for LLMs that integrates captions, summaries, and similar QA into a cohesive instructional framework. The template of the prompt sample is:

\begin{center}
\begin{tabular}{|p{7.5cm}|}\hline
\textbf{Instruction text:} Please reason the answers to the questions according to the contexts. $\backslash n$ \\
\textbf{Contexts:} Rerank\_Caption: \textcolor{blue}{$e$}$\backslash n+$ Summary: \textcolor{blue}{$s$}$\backslash n+$ Similar\_Question: \textcolor{blue}{$q^i$}$\backslash n+$ Similar\_Answer: \textcolor{blue}{$a^i$}$\backslash n + $ Question: \textcolor{blue}{$Q$}$\backslash n + $ Answer: \\ \hline
\end{tabular}
\end{center}

Specifically, the guidance provided to the LLMs is explicitly defined: “\textit{Please infer the answer to the question based on the context}”. This directive primes the LLMs for context-based reasoning. The structured input to the LLMs is formatted as follows: the initial section labeled “\textit{Contexts}” encapsulates all the captions and the summary, providing a rich narrative backdrop. This is followed by a series of QA examples, formatted as “\textit{Question: [question] Answer: [answer]}”, which are sequentially concatenated to build a narrative flow and context reinforcement. The prompt culminates with the presentation of the current question, formatted distinctly as “\textit{Question: [current question] Answer:}”, which sets the stage for the LLM’s response generation. The last question is positioned to prompt LLMs to provide a correct answer, focusing solely on generating a new answer.

\section{Experiment}
\label{sec:4}
In this section, a series of experiments using two publicly available datasets are conducted to address the following  
questions: \textbf{RQ1:} How does the performance of SCRA-VQA compare to other VQA models?  \textbf{RQ2:} How do the main modules influence the performance of SCRA-VQA? \textbf{RQ3:} How do key hyperparameters affect SCRA-VQA performance? \textbf{RQ4:} How does SCRA-VQA perform when tested on real-word datasets?

\subsection{Environment Setup}
\textbf{Datasets.} This paper selects two knowledge-based VQA datasets to evaluate our method: 1) The OK-VQA \cite{11} test set contains 5,046 questions requiring external knowledge resources. 2) A-OKVQA \cite{29} is an augmented dataset built on the OK-VQA dataset and requires a broader range of knowledge types compared to OK-VQA. Additionally, most questions in the VQAv2 \cite{30} dataset focus on straightforward visual recognition tasks, whereas our method emphasizes tasks requiring external knowledge or logical reasoning. Therefore, we did not test our method on VQAv2.

\noindent\textbf{Experimental Details.} This paper uses BLIP with 446M parameters for image captions to generate captions for the regions matched between the image and the question. To determine the relevant regions between the image and the question, we use the cross-attention layers of the image-based text encoder in BLIP to generate GradCAM and then sample K=20 image segments from these identified regions. We primarily use the open-source OPT \cite{31} model for the LLMs. In our experiments, we also tested various other open-source models to further demonstrate our method's strong compatibility. These models can all be experimented on using 2$\times$Nvidia RTX4090, which is generally affordable for most research labs.

\noindent\textbf{Baseline Methods.} This paper categorizes previous methods into four types: 1) Zero-shot methods with frozen LLMs, which utilize frozen LLMs, including PICa \cite{5}, PNP-VQA \cite{32}, and Img2Promp \cite{10}. 2) Zero-shot methods with extra multi-modal pretraining, which require large-scale vision-language datasets, including VL-T5 \cite{33}, FewVLM \cite{34}, VLKD \cite{35}, and Frozen \cite{36}. 3) Zero-shot Evaluation with Access to Answer Candidates, including VQ2A \cite{9} and WeaQA \cite{37}. It is worth noting that these methods assume the presence of scenarios where answer candidates may not be available in practice, thus requiring careful comparison. 4) Few-Shot Evaluation, which involves evaluation with a few samples, including PICa, FewVLM, and ClipCap \cite{38}. 
\begin{table}[!t]  
\centering  
\caption{Results on OK-VQA and A-OKVQA. $\times$ denotes methods that do not require end-to-end training, while $\checkmark$ indicates those that do. Methods without end-to-end training but assuming access to training samples are marked with $\dagger$.}  
\begin{tabular}{p{3.2cm}ccccc}  
    \toprule       
    \makecell[l]{\textbf{Methods}} & \textbf{Parameters} & \textbf{\begin{tabular}[c]{@{}c@{}}End-to-End\\ Training\end{tabular}} & \textbf{\begin{tabular}[c]{@{}c@{}}Shot\\ Number\end{tabular}} &\textbf{\begin{tabular}[c]{@{}c@{}}OK-VQA\\ test\end{tabular}} &\textbf{\begin{tabular}[c]{@{}c@{}}A-OKVQA\\ val.\end{tabular}} \\  
    \midrule   
    \multicolumn{6}{l}{\textbf{\textit{Zero-Shot Evaluation with Extra End-to-End Training}}} \\  
    \makecell[l]{VL-T5$_{no-vqa}$ \cite{33}}& 288M & $\checkmark$ & 0 & 5.8 & - \\  
    \makecell[l]{FewVLM$_{base}$ \cite{34}}& 288M & $\checkmark$ & 0 & 11.6 & - \\  
    \makecell[l]{FewVLM$_{large}$ \cite{34}}& 804M & $\checkmark$ & 0 & 16.5 & - \\  
    \makecell[l]{VLKD$_{ViT-B/16}$ \cite{35}}& 494M & $\checkmark$ & 0 & 10.5 & - \\  
    \makecell[l]{VLKD$_{ViT-L/14}$ \cite{35}} & 713M & $\checkmark$ & 0 & 13.3 & - \\  
   \makecell[l]{Frozen$_{7B}$ \cite{36}}& 7B & $\checkmark$ & 0 & 5.9 & - \\  
    \hline  
    \multicolumn{6}{l}{\textbf{\textit{Zero-shot Evaluation with Access to Answer Candidates}}} \\  
    \makecell[l]{WeaQA-ZSL \cite{37}}& - & $\checkmark$ & 0 & 15.8 & - \\  
    \makecell[l]{VQ2A \cite{9}}& - & $\checkmark$ & 0 & 19.8 & 21.5 \\  
    \hline  
    \multicolumn{6}{l}{\textbf{\textit{Few-Shot Evaluation}}} \\  
    \makecell[l]{ClipCap$_{Cap}$ \cite{38}}& 175B & $\times$ & 10 & 10.0 & 16.6 \\  
    \makecell[l]{ClipCap$_{Rel}$ \cite{38}}& 175B & $\times$ & 10 & - & 18.1 \\  
    \makecell[l]{FewVLM$_{base}$ \cite{34}}& 288M & $\checkmark$ & 16 & 15.0 & - \\  
    \makecell[l]{FewVLM$_{large}$ \cite{34}}& 804M & $\checkmark$ & 16 & 23.1 & - \\  
    \makecell[l]{PICA$_{175B\dagger}$ \cite{5}}& 175B & $\times$ & 1 & 36.4 & - \\  
    \hline
    \multicolumn{6}{l}{\textbf{\textit{Zero-Shot Evaluation with Frozen Large Language Model}}} \\  
   \makecell[l]{PICA$_{175B}$ \cite{5}}& 175B & $\times$ & 0 & 17.7 & - \\  
    \makecell[l]{PNP-VQA$_{3B}$ \cite{35}}& 3.9B & $\times$ & 0 & 34.1 & 33.4 \\  
    \makecell[l]{PNP-VQA$_{11B}$ \cite{35}}& 11.9B & $\times$ & 0 & 35.9 & 33.8 \\  
    \makecell[l]{Img2Prompt$_{7B}$ \cite{10}}& 8.3B & $\times$ & 0 & 38.2 & 33.3 \\  
    \makecell[l]{\textbf{SCRA-VQA$_{6.7B}$} }& 8.3B & $\times$ & 0 & \textbf{38.8} & \textbf{34.6} \\ 
    \bottomrule
\end{tabular}  
\label{table:2}  
\end{table} 
\subsection{Main Results (RQ1)}
\textbf{Comparing the performance of Zero-Shot and Few-Shot models.} As shown in Table~\ref{table:2}, SCRA-VQA outperforms PICa on OK-VQA, a strong zero-shot baseline without requiring large-scale end-to-end training, and enhances performance on the OK-VQA dataset by 8.1\% compared to the PNP-VQA$_{11B}$ model while maintaining a lightweight structure. Against Img2Prompt of similar size, SCRA-VQA scores 38.8\% on OK-VQA and 34.6\% on A-OKVQA, surpassing Img2Prompt's 38.2\% and 33.3\%, respectively. These results confirm the effectiveness of our method, especially on A-OKVQA, which requires more sophisticated reasoning, highlighting its capability to handle complex queries. In the context of end-to-end pre-training and few-shot models, SCRA-VQA remains competitive, achieving more than twice the results of ClipCap on A-OKVQA. Despite the lack of large-scale vision-language (V-L) pretraining, SCRA-VQA demonstrates commendable performance when compared to methods that rely on such extensive pretraining. Overall, SCRA-VQA achieves robust performance without the necessity of extensive pretraining, positioning it as a highly effective approach for a variety of scenarios.

\subsection{Ablation Study (RQ2)}
\textbf{Effect of Different Prompting Content.} Table~\ref{table:3} demonstrates the performance of the A-OKVQA and OK-VQA datasets under different prompting contents. We observed a significant increase in accuracy when the Prompt Template included Summary and Caption\_Rerank. Experimental results indicate their role in effectively connecting the dots between language model pretraining and VQA tasks, addressing modality disconnection issues. By enriching the prompts, we can better leverage the capabilities of LLMs, facilitating a more seamless integration across different modalities in VQA scenarios.

\begin{table}[!t]
\centering
\caption{Ablation of different prompt template content. It includes Instruction (I), Caption (C), Question-Answer Pairs (QAP), Summary (S), and Rerank Caption (RC). The results are run with OPT6.7B.}
\begin{tabular}{lcccc}
\toprule
\textbf{Prompt Template} & \textbf{Summary} & \textbf{Rerank} & \textbf{OK-VQA} & \textbf{A-OKVQA} \\ \midrule
\makecell[l]{I} & $\times$ & $\times$ & 15.21 & 2.33 \\ \midrule
\makecell[l]{I + C} & $\times$ & $\times$ & 35.32 & 29.50 \\ \midrule
\makecell[l]{I + C + QAP} & $\times$ & $\times$ & 38.29 & 33.32 \\ \midrule
\makecell[l]{I +  S + QAP} & $\checkmark$ & $\times$ & 38.00 & 33.42 \\ \midrule
\makecell[l]{I + C + S + QAP} & $\checkmark$ & $\times$ & 38.10 & 33.10 \\ \midrule
\makecell[l]{I + RC + S + QAP} & $\checkmark$ & $\checkmark$ & \textbf{38.82} & \textbf{34.61} \\ \bottomrule
\end{tabular}
\label{table:3}
\end{table}

\begin{table}[!t]
\centering
\caption{Ablations on prompts format. Repetition indicates whether it needs to be used more than once.}
\begin{tabular}{cccc}
\toprule
\makecell[l]{\textbf{Methods}} & \textbf{Repetition} & \textbf{OK-VQA} & \textbf{A-OKVQA}  \\ \midrule
\makecell[l]{S+C+QA}&$\checkmark$ & 36.5 & 32.2 \\ 
\makecell[l]{C+QA+S}&$\checkmark$ & 36.3 & 32.5 \\ 
\makecell[l]{MC+MQA+S}&$\times$ & 37.7 & 32.9 \\ 
\makecell[l]{S+MC+MQA}&$\times$& 38.0 & 33.9 \\ 
\makecell[l]{MC+S+MQA}&$\times$& \textbf{38.8} & \textbf{34.6} \\ \bottomrule
\end{tabular}
\label{table:4}
\end{table}

\noindent\textbf{Effect of Prompt Text Design.} We explore several methods for constructing prompts for LLMs. The first method appends a QA pair after each caption, with the summary at either the beginning or the end. This is described as Summary Caption Question-Answer (S+C+QA) and Caption Question-Answer Summary (C+QA+S). This type of prompt must be repeated repeatedly to display all the information. The second method presents all the captions followed by all the QA pairs and then adds the summary. This is represented as MultiCaption MultiQA Summary (MC+MQA+S), Summary MultiCaption MultiQA (S+MC+MQA), and MultiCaption Summary MultiQA (MC+S+MQA). Experiments (as shown in Table~\ref{table:4}) reveal that the second method significantly outperforms the first in performance. This result might be due to the first design. It could cause the LLM to overly rely on the caption content while neglecting other potentially significant or supplementary information sources.

\subsection{Hyper-parameter Sensitivity (RQ3)}
\begin{figure}[!t]
\centering
\includegraphics[scale=0.2]{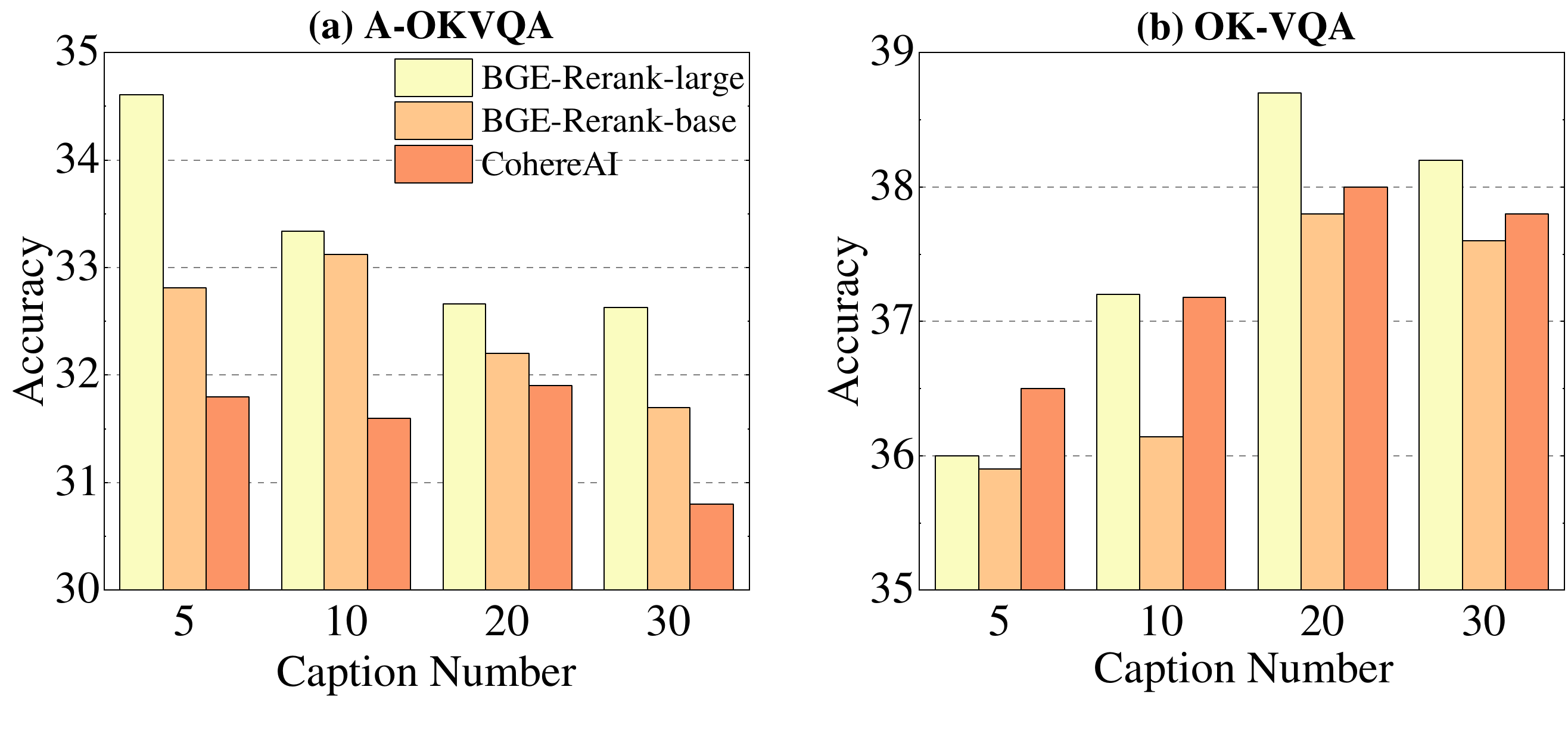} 
\caption{The number of captions in the A-OKVQA and OK-VQA datasets was used for hyperparameter analysis.}
\label{figure:4}
\end{figure}

\textbf{Effect of Caption Quantity and Quality}: Fig.~\ref{figure:4} (a) and (b) illustrate the performance on the OK-VQA and A-OKVQA datasets using different captions. Here, we utilize three mainstream rerank models: BGE-Rerank-large, BGE-Rerank-base, and CohereAI. The figures indicate that the BGE-Rerank-large model performs the best on both datasets. Additionally, BGE-Rerank-large achieves the best performance on the OK-VQA dataset when the number of captions is 20. Conversely, on the A-OKVQA dataset, the model performs best with five captions. Increasing the number of captions beyond this point results in a drop in performance, indicating that a certain number of captions causes LLMs to ignore important information.

\begin{figure}[!t]
\centering
\includegraphics[scale=0.45]{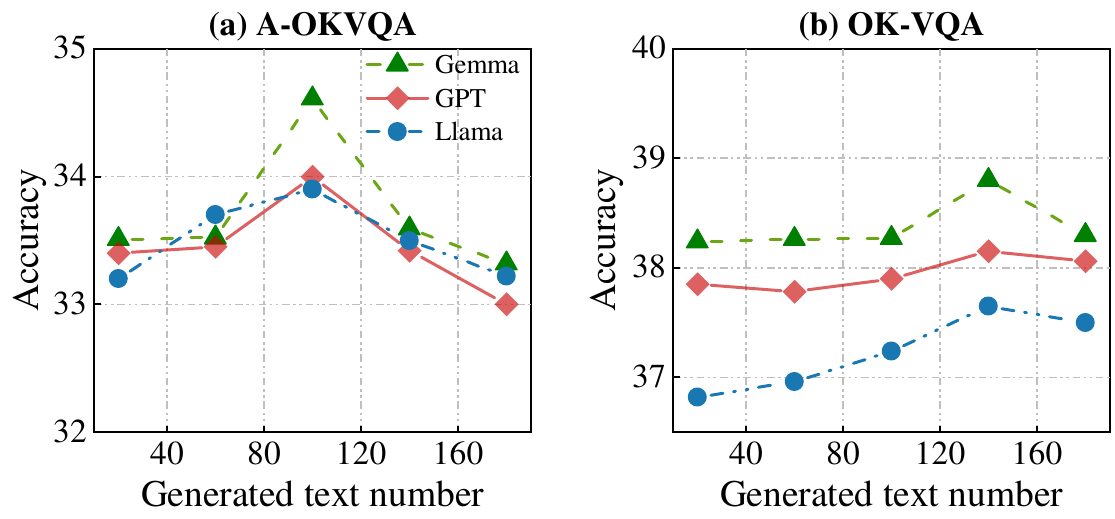}
\caption{Hyper-parameter analysis for different generated text lengths.}
\label{figure:5}
\end{figure}

\noindent\textbf{Effect of Summary Text Length and Quality}: Given the input limitations of LLMs, the length and quality of the summary text can affect their answer generation.  To explore this effect, we conduct experiments using Gemma, GPT, and Llama. As shown in Fig.~\ref{figure:5}, Gemma performs best on the A-OKVQA dataset with a text length of 100, while optimal performance on the OK-VQA dataset is achieved with a text length of 140. This difference is due to the varying number of captions in the datasets. Performance declines for almost all models when text length increases to 160, as longer texts tend to include more irrelevant information.

\begin{figure}[t]
\centering  
\includegraphics[scale=0.60]{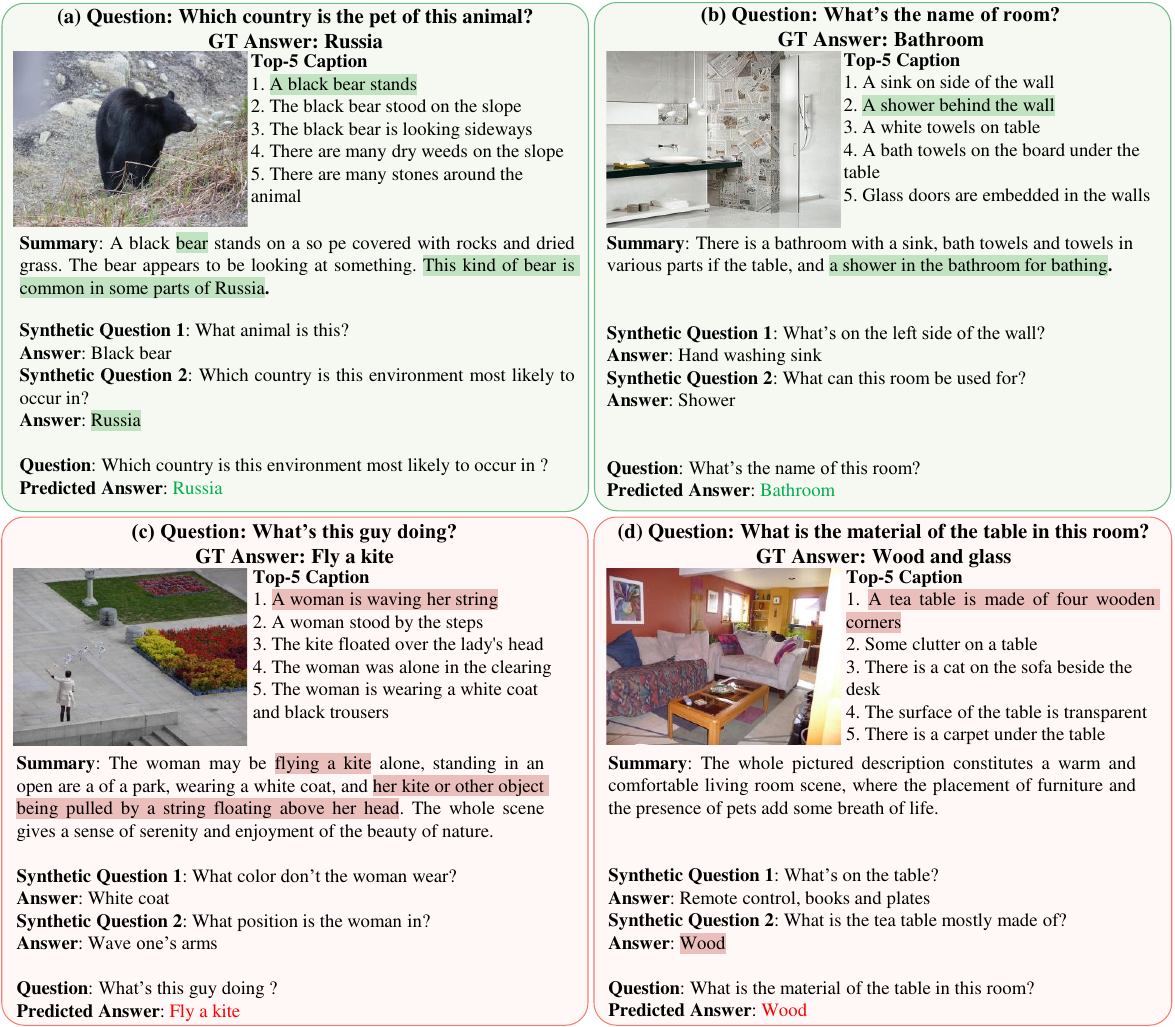} 
\caption{Visualization of SCRA-VQA predicted answers and prompt information. Green and Red refer to correct and incorrect results, respectively.}
\label{figure:6}
\end{figure}

\subsection{Case Study (RQ4)}
In Fig.~\ref{figure:6}, we present four examples, including both successful and failed cases. For example, in Fig.~\ref{figure:6}(a), the reranked captions all contain the necessary information, like “shower” and “bathroom”. The summarized text organizes caption information, and the similar sub-question-answer pairs provide ample contextual learning for the LLMs. However, errors in the image-to-text conversion process can negatively impact subsequent modules. For example, in Fig.~\ref{figure:6}(d), BLIP incorrectly identifies a small kite as a string, leading to inadequate information for the question. These results highlight the necessity of appropriate commonsense reasoning in open-ended VQA tasks.

\section{Conclusion}
\label{sec:5}
This paper introduces SCRA-VQA, designed to harness existing LLMs' knowledge and reasoning capabilities for zero-shot VQA tasks. Specifically, SCRA-VQA extracts and summarizes image information, providing ample context to supply LLMs with sufficient visual information and task guidance. This method eliminates the need for costly end-to-end visual-language alignment. Experimental results demonstrate that SCRA-VQA enables various LLMs to achieve zero-shot VQA performance comparable to or surpass other methods requiring expensive end-to-end training. Future research could explore refining SCRA-VQA’s information extraction techniques, extending its applicability beyond visual data to diverse modalities, and enhancing methods to integrate multi-modal knowledge into LLMs.

\subsubsection{\ackname} This work was supported by the National Natural Science Foundation of China (No. 62207011, 62377009, 62407013, 62101179), the Open Fund of Hubei Key Laboratory of Big Data Intelligent Analysis and Application, Hubei University (No. 2024BDIAA05), the Open Fund of Key Laboratory of Intelligent Sensing System and Security of Hubei University, Ministry of Education (No. KLISSS202410), the Science and Technology Support Plan for Youth Innovation of Colleges and Universities of Shandong Province of China (NO. 2023KJ370), the Key Research and Development Project of Hubei Province (NO. 2021BAA184), the Major Program (JD) of Hubei Province (NO. 2023BAA018).


%
%
%
%
%

\begin{thebibliography}{8}
\bibitem{1}
Lu, S., Liu, M., Yin, L., Yin, Z., Liu, X., Zheng, W.: The multi-modal fusion in visual question answering: a review of attention mechanisms. PeerJ Comput. Sci. 9, e1400 (2023)

\bibitem{2}
Wu, Q., Teney, D., Wang, P., Shen, C., Dick, A.R., van den Hengel, A.: Visual question answering: A survey of methods and datasets. Comput. Vis. Image Underst. 163, 21--40 (2017)


\bibitem{3}
Sood, E., Kögel, F., Müller, P., Thomas, D., Băce, M., Bulling, A.: Multimodal Integration of Human-Like Attention in Visual Question Answering. In: Proceedings of the IEEE/CVF Conference on Computer Vision and Pattern Recognition, pp. 2648--2658. (2023)

\bibitem{4}
Cadène, R., Ben-Younes, H., Cord, M., Thome, N.: MUREL: Multimodal Relational Reasoning for Visual Question Answering. In: Proceedings of the IEEE Conference on Computer Vision and Pattern Recognition, pp. 1989--1998. (2019)



\bibitem{5}
Yang, Z., Gan, Z., Wang, J., Hu, X., Lu, Y., Liu, Z., Wang, L.: An Empirical Study of GPT-3 for Few-Shot Knowledge-Based VQA. In: Proceedings of the Thirty-Sixth AAAI Conference on Artificial Intelligence, pp. 3081--3089. (2022)

\bibitem{6}
Heo, Y.-J., Kim, E.-S., Choi, W.S., Zhang, B.-T.: Hypergraph Transformer: Weakly-Supervised Multi-hop Reasoning for Knowledge-based Visual Question Answering. In: Proceedings of the 60th Annual Meeting of the Association for Computational Linguistics, pp. 373--390. (2022)


\bibitem{7}
Shao, Z., Yu, Z., Wang, M., Yu, J.: Prompting Large Language Models with Answer Heuristics for Knowledge-Based Visual Question Answering. In: Proceedings of the IEEE/CVF Conference on Computer Vision and Pattern Recognition, pp. 14974--14983. (2023)

\bibitem{8}
Li, J., Li, D., Xiong, C., Hoi, S.C.H.: BLIP: Bootstrapping Language-Image Pre-training for Unified Vision-Language Understanding and Generation. In: Proceedings of the International Conference on Machine Learning, vol. 162, pp. 12888--12900. (2022)

\bibitem{9}
Changpinyo, S., Kukliansky, D., Szpektor, I., Chen, X., Ding, N., Soricut, R.: All You May Need for VQA are Image Captions. In: Proceedings of the 2022 Conference of the Association for Computational Linguistics, pp. 1947--1963. (2022)

\bibitem{10}
Guo, J., Li, J., Li, D., Tiong, A.M.H., Li, B., Tao, D., Hoi, S.C.H.: From Images to Textual Prompts: Zero-shot Visual Question Answering with Frozen Large Language Models. In: Proceedings of the IEEE/CVF Conference on Computer Vision and Pattern Recognition, pp. 10867--10877. (2023)

\bibitem{11}
Marino, K., Rastegari, M., Farhadi, A., Mottaghi, R.: OK-VQA: A Visual Question Answering Benchmark Requiring External Knowledge. In: Proceedings of the IEEE Conference on Computer Vision and Pattern Recognition, pp. 3195--3204. (2019)

\bibitem{12}
Jiang, H., Misra, I., Rohrbach, M., Learned-Miller, E.G., Chen, X.: In Defense of Grid Features for Visual Question Answering. In: Proceedings of the 2020 IEEE/CVF Conference on Computer Vision and Pattern Recognition, pp. 10264--10273. (2020)

\bibitem{13}
Shen, S., Li, L.H., Tan, H., Bansal, M., Rohrbach, A., Chang, K.-W., Yao, Z., Keutzer, K.: How Much Can CLIP Benefit Vision-and-Language Tasks? In: Proceedings of the Tenth International Conference on Learning Representations. (2022)

\bibitem{14}
Zhang, P., Li, X., Hu, X., Yang, J., Zhang, L., Wang, L., Choi, Y., Gao, J.: VinVL: Revisiting Visual Representations in Vision-Language Models. In: Proceedings of the IEEE Conference on Computer Vision and Pattern Recognition, pp. 5579--5588. (2021)

\bibitem{15}
Hu, R., Andreas, J., Rohrbach, M., Darrell, T., Saenko, K.: Learning to Reason: End-to-End Module Networks for Visual Question Answering. In: Proceedings of the IEEE International Conference on Computer Vision, pp. 804--813. (2017)

\bibitem{16}
Gan, C., Li, Y., Li, H., Sun, C., Gong, B.: VQS: Linking Segmentations to Questions and Answers for Supervised Attention in VQA and Question-Focused Semantic Segmentation. In: Proceedings of the IEEE International Conference on Computer Vision, pp. 1829--1838. (2017)

\bibitem{17}
Yi, K., Wu, J., Gan, C., Torralba, A., Kohli, P., Tenenbaum, J.: Neural-Symbolic VQA: Disentangling Reasoning from Vision and Language Understanding. In: Advances in Neural Information Processing Systems, pp. 1039--1050. (2018)

\bibitem{18}
Barbiero, P., Ciravegna, G., Giannini, F., Espinosa Zarlenga, M., Magister, L.C., Tonda, A., Lio, P., Precioso, F., Jamnik, M., Marra, G.: Interpretable Neural-Symbolic Concept Reasoning. In: Proceedings of the International Conference on Machine Learning, pp. 1801--1825. (2023)

\bibitem{19}
Wu, J., Mooney, R.J.: Entity-Focused Dense Passage Retrieval for Outside-Knowledge Visual Question Answering. In: Proceedings of the 2022 Conference on Empirical Methods in Natural Language Processing, pp. 8061--8072. (2022)

\bibitem{20}
Zhang, L., Liu, S., Liu, D., Zeng, P., Li, X., Song, J., Gao, L.: Rich Visual Knowledge-Based Augmentation Network for Visual Question Answering. IEEE Trans. Neural Networks Learn. Syst. 32(10), 4362--4373 (2021)

\bibitem{21}
Lin, Y., Xie, Y., Chen, D., Xu, Y., Zhu, C., Yuan, L.: REVIVE: Regional Visual Representation Matters in Knowledge-Based Visual Question Answering. In: Advances in Neural Information Processing Systems (2022)

\bibitem{22}
Narasimhan, M., Lazebnik, S., Schwing, A.G.: Out of the Box: Reasoning with Graph Convolution Nets for Factual Visual Question Answering. In: Advances in Neural Information Processing Systems, pp. 2659--2670 (2018)

\bibitem{23}
Shah, S., Mishra, A., Yadati, N., Talukdar, P.P.: KVQA: Knowledge-Aware Visual Question Answering. In: Proceedings of The Thirty-Third AAAI Conference on Artificial Intelligence, pp. 8876--8884 (2019)

\bibitem{24}
Gui, L., Wang, B., Huang, Q., Hauptmann, A., Bisk, Y., Gao, J.: KAT: A Knowledge Augmented Transformer for Vision-and-Language. In: Proceedings of the 2022 Conference of the North American Chapter of the Association for Computational Linguistics, pp. 956--968 (2022)

\bibitem{25}
Kan, B., Wang, T., Lu, W., Zhen, X., Guan, W., Zheng, F.: Knowledge-Aware Prompt Tuning for Generalizable Vision-Language Models. In: Proceedings of the IEEE/CVF International Conference on Computer Vision, pp. 15624--15634 (2023)

\bibitem{26}
Selvaraju, R.R., Cogswell, M., Das, A., Vedantam, R., Parikh, D., Batra, D.: Grad-CAM: Visual Explanations from Deep Networks via Gradient-Based Localization. Int. J. Comput. Vis. 128(2), 336--359 (2020)


\bibitem{27}
Schick, T., Schütze, H.: Exploiting Cloze-Questions for Few-Shot Text Classification and Natural Language Inference. In: Proceedings of the 16th Conference of the European Chapter of the Association for Computational Linguistics, pp. 255--269 (2021)

\bibitem{28}
Alayrac, J.-B., Donahue, J., Luc, P., Miech, A., Barr, I., Hasson, Y., Lenc, K., Mensch, A., Millican, K., Reynolds, M., Ring, R., Rutherford, E., Cabi, S., Han, T., Gong, Z., Samangooei, S., Monteiro, M., Menick, J.L., Borgeaud, S., Brock, A., Nematzadeh, A., Sharifzadeh, S., Binkowski, M., Barreira, R., Vinyals, O., Zisserman, A., Simonyan, K.: Flamingo: A Visual Language Model for Few-Shot Learning. In: Advances in Neural Information Processing Systems (2022)

\bibitem{29}
Schwenk, D., Khandelwal, A., Clark, C., Marino, K., Mottaghi, R.: A-OKVQA: A Benchmark for Visual Question Answering Using World Knowledge. In: Proceedings of the 17th European Conference on Computer Vision, pp. 146--162 (2022)

\bibitem{30}
Goyal, Y., Khot, T., Summers-Stay, D., Batra, D., Parikh, D.: Making the V in VQA Matter: Elevating the Role of Image Understanding in Visual Question Answering. In: Proceedings of the 2017 IEEE Conference on Computer Vision and Pattern Recognition, pp. 6325--6334 (2017)

\bibitem{31}
Zhang, S., Roller, S., Goyal, N., Artetxe, M., Chen, M., Chen, S., Dewan, C., Diab, M.T., Li, X., Lin, X.V., Mihaylov, T., Ott, M., Shleifer, S., Shuster, K., Simig, D., Singh Koura, P., Sridhar, A., Wang, T., Zettlemoyer, L.: OPT: Open Pre-trained Transformer Language Models. CoRR abs/2205.01068 (2022)

\bibitem{32}
Tiong, A.M.H., Li, J., Li, B., Savarese, S., Hoi, S.C.H.: Plug-and-Play VQA: Zero-shot VQA by Conjoining Large Pretrained Models with Zero Training. In: Proceedings of the Association for Computational Linguistics, pp. 951--967 (2022)

\bibitem{33}
Cho, J., Lei, J., Tan, H., Bansal, M.: Unifying Vision-and-Language Tasks via Text Generation. In: Proceedings of the 38th International Conference on Machine Learning, vol. 139, pp. 1931--1942 (2021)

\bibitem{34}
Jin, W., Cheng, Y., Shen, Y., Chen, W., Ren, X.: A Good Prompt Is Worth Millions of Parameters: Low-resource Prompt-based Learning for Vision-Language Models. In: Proceedings of the 60th Annual Meeting of the Association for Computational Linguistics, pp. 2763--2775 (2022)

\bibitem{35}
Dai, W., Hou, L., Shang, L., Jiang, X., Liu, Q., Fung, P.: Enabling Multimodal Generation on CLIP via Vision-Language Knowledge Distillation. In: Proceedings of the Association for Computational Linguistics, pp. 2383--2395 (2022)

\bibitem{36}
Tsimpoukelli, M., Menick, J., Cabi, S., Eslami, S.M.A., Vinyals, O., Hill, F.: Multimodal Few-Shot Learning with Frozen Language Models. In: Advances in Neural Information Processing Systems, pp. 200--212 (2021)

\bibitem{37}
Banerjee, P., Gokhale, T., Yang, Y., Baral, C.: WeaQA: Weak Supervision via Captions for Visual Question Answering. In: Proceedings of the Association for Computational Linguistics, pp. 3420--3435 (2021)

\bibitem{38}
Mokady, R., Hertz, A., Berman, A.H.: ClipCap: CLIP Prefix for Image Captioning. CoRR abs/2111.09734 (2021)

\end{thebibliography}
%

\end{document}